%% file: ms.tex
\documentclass[letterpaper, 10 pt, conference]{./ieeeconf}  

\IEEEoverridecommandlockouts                              

\usepackage{graphicx}
\graphicspath{ {./imgs/} }
\usepackage{amsmath} 
\usepackage{amssymb}  

\usepackage[style=ieee,natbib=true,maxcitenames=1,mincitenames=1]{biblatex} 
\usepackage{balance}
\usepackage{adjustbox}
\usepackage{hyperref}
\usepackage{url}
\usepackage{tablefootnote}

\usepackage{xcolor}
\usepackage{comment}
\usepackage{subfiles}
\usepackage{makecell}
\usepackage{multirow}
\usepackage{bm}

\newcommand{\editold}[1]{\textcolor{black}{#1}}

\newcommand{\vect}[1]{\mathbf{\bm{#1}}}
\newcommand{\mk}[1]{\mkern-#1mu}

\newcommand{\W}{W\mk3}  
\newcommand{\I}{I\mk3}  
\newcommand{\B}{B\mk3}  
\newcommand{\A}{A\mk3}  

\usepackage{subfig}

\title{\LARGE \bf
Extrinsic Calibration of Multiple Inertial Sensors\\ from \editold{Arbitrary Trajectories}
}
\author{
Jongwon Lee$^{1}$, David Hanley$^{2}$, and Timothy Bretl$^{1}$
\thanks{$^{1}$Jongwon Lee and Timothy Bretl are with the Department of Aerospace Engineering, University of Illinois at Urbana-Champaign, Urbana, IL 61801, USA (Email: \texttt{\{jongwon5, tbretl\}@illinois.edu}).}
\thanks{$^{2}$David Hanley is with the Department of Electrical and Computer Engineering, University of Illinois at Urbana-Champaign, Urbana, IL 61801, USA (Email: {\tt\small hanley6@illinois.edu}).}
}

\begin{document}

\maketitle
\thispagestyle{empty}
\pagestyle{empty}

\subfile{sections/abstract.tex}

\section{Introduction}

\subfile{sections/introduction.tex}

\label{section:Introduction}

\section{Methodology}

\subfile{sections/methodology.tex}

\label{section:Methodology}

\section{Simulation Experiments}

\subfile{sections/simulation-experiments.tex}

\label{section:Simulation Experiments}

\section{Hardware Experiments}

\subfile{sections/real-world-experiments.tex}

\label{section:Real-World Experiments}

\section{Conclusion}

\subfile{sections/conclusion.tex}

\label{section:Conclusion}

\section*{ACKNOWLEDGEMENT}
This work was supported by the NASA Grant No. STTR-80NSSC20C0020.

\balance
\printbibliography

\end{document}

%% file: sections/abstract.tex

\begin{abstract}

We present a method of extrinsic calibration for a system of multiple inertial measurement units (IMUs) that estimates the relative pose of each IMU on a rigid body using only measurements from the IMUs themselves\editold{, without the need to prescribe the trajectory}. Our method is based on solving a nonlinear least-squares problem that penalizes inconsistency between measurements from pairs of IMUs. We validate our method with experiments both in simulation and in hardware. In particular, we show that it meets or exceeds the performance---in terms of error, success rate, and computation time---of an existing, state-of-the-art method that does not rely only on IMU measurements and instead requires the use of a camera and a fiducial marker. We also show that the performance of our method is largely insensitive to the choice of trajectory along which IMU measurements are collected.

\end{abstract}

%% file: sections/introduction.tex
Inertial sensors are widely used in mobile robot navigation systems.
These sensors are typically packaged in an inertial measurement unit (IMU), which consists of a triaxial accelerometer to measure specific force along orthogonal axes and of three gyroscopes to measure angular rates about these same axes.
While it is still common to use only a single IMU as the basis for a navigation system, it has been recognized that the simultaneous use of multiple IMUs on the same robot may result in higher measurement accuracy, increased bandwidth, and better fault tolerance than could have been achieved with a single IMU of the same total size, weight, power, and cost~\cite{nilsson2016inertial, skog2016inertial, parsa2005estimation, wijayasinghe2018study, zhang2020lightweight, eckenhoff2019sensor, eckenhoff2020mimc}.
In order to realize these benefits, however, it is necessary to perform {\em extrinsic calibration}---that is, to estimate the relative position and orientation (i.e., the relative {\em pose}) of each IMU with respect to the robot on which they are all mounted.

Existing methods of extrinsic calibration for multi-IMU systems fall into three categories.
First, there are methods that rely on the use of instruments (e.g., rate tables) to precisely control the trajectory of the robot on which the IMUs are mounted~\cite{cho2005calibration, schopp2010design, he2013novel, nilsson2014aligning}. Calibration is then a matter of choosing the relative pose of each IMU that minimizes the difference between expected and actual measurements.
We largely ignore methods in this first category, since our goal in this paper is to perform extrinsic calibration from data collected in flight along arbitrary trajectories.
Second, there are methods that rely on the use of aiding sensors, for example a camera.
``Kalibr'' is one such method---it assumes the navigation system consists of a single camera and of multiple IMUs, and applies well-known algorithms to estimate the pose of each IMU (one-by-one) with respect to the camera, given images of fiducial markers~\cite{furgale2013unified, rehder2016extending}.
Again, we largely ignore methods in this second category, since our goal in this paper is to perform extrinsic calibration without the use of aiding sensors.
However, we do use Kalibr as a benchmark in our hardware experiments (Section~\ref{section:Real-World Experiments}), since it is available as open-source and provides excellent performance.
Third, there are methods that---like ours---require neither instruments nor aiding sensors~\cite{qiu2020real, kim2017line, schopp2016self, kortier2019use}. Instead, these methods require only measurements from the IMUs themselves. We were particularly inspired by the method of \citeauthor{schopp2016self}~\cite{schopp2016self}, which---like ours---is based on solving a nonlinear least-squares problem that penalizes inconsistency between expected and actual measurements.
Indeed, one way to think about what we are trying to do in this paper is to extend methods in this third category to eliminate certain limitations (e.g., applying only to arrays of accelerometers and not gyroscopes) and to relax certain assumptions (e.g., that accelerometer and gyroscope axes are perfectly aligned in each IMU).

Section~\ref{section:Methodology} presents our method of extrinsic calibration.
Section~\ref{section:Simulation Experiments} validates our method with experimental results in simulation using OpenVINS~\cite{geneva2020openvins}.
Section~\ref{section:Real-World Experiments} validates our method with experimental results in hardware, using Kalibr as a benchmark.
Section~\ref{section:Conclusion} discusses a number of directions for future work.

Both the code and dataset used in this paper are available online.\footnote{\url{https://github.com/jongwonjlee/mix-cal}}

\renewcommand{\arraystretch}{1.5}

\begin{table}
\begin{center}
\captionsetup{justification=centering}
\caption{\textsc{Comparing different methods of \\extrinsic calibration for multiple IMUs}}
\label{table:comparison}
\begin{adjustbox}{width=.45\textwidth}
\begin{tabular}{|c|ccc|>{\centering\arraybackslash}p{0.3cm}>{\centering\arraybackslash}p{0.3cm}>{\centering\arraybackslash}p{0.3cm}|}
    \Xhline{\arrayrulewidth}
    & \multicolumn{3}{c|}{\thead{characteristics}} & \multicolumn{3}{c|}{\thead{estimated\\ quantities}} \\
    method & \makecell{free from \\camera?} & \makecell{arbitrary \\trajectory?} & \makecell{code \\available?} & $\vect{p}$ & $\vect{q}$ & $^{g}_{I}\vect{q}$ \\
    \Xhline{\arrayrulewidth}
    \citeauthor{qiu2020real}~\cite{qiu2020real}                 & $\bullet$ & $\bullet$ & -         & -         & $\bullet$ & -         \\
    \citeauthor{kim2017line}~\cite{kim2017line}                 & $\bullet$ & $\bullet$ & -         & $\bullet$ & $\bullet$ & -         \\
    \citeauthor{schopp2016self}~\cite{schopp2016self}           & $\bullet$ & $\bullet$ & -         & $\bullet$ & $\bullet$ & -         \\
    \citeauthor{rehder2016extending}~\cite{rehder2016extending} & -         & -         & $\bullet$ & $\bullet$ & $\bullet$ & $\bullet$ \\
    \citeauthor{kortier2019use}~\cite{kortier2019use}           & $\bullet$ & -         & -         & $\bullet$ & $\bullet$ & $\bullet$ \\
    \textbf{Our Method}                                         & $\bullet$ & $\bullet$ & $\bullet$ & $\bullet$ & $\bullet$ & $\bullet$ \\
    \Xhline{\arrayrulewidth}
\end{tabular}
\end{adjustbox}
\captionsetup{width=.95\textwidth}
\caption*{\footnotesize $\vect{p}$: IMU position, $\vect{q}$: IMU orientation, $^{g}_{I}\vect{q}$: gyroscope misalignment, \\$\bullet$: applicable, -: not applicable}
\end{center}
\end{table}

\renewcommand{\arraystretch}{1.0}

%% file: sections/methodology.tex
\subsection{Notation}

We use ${}^{\B}_{\A}\vect{R}\:\in\:SO(3)$ to denote the rotation matrix that describes the orientation of frame $\mathcal{F}_A$ in the coordinates of frame $\mathcal{F}_B$.
We use $\vect{C}(\cdot)$ to denote the operator that converts a unit quaternion ${}^{\B}_{\A}\vect{q}$ to the corresponding rotation matrix ${}^{\B}_{\A}\vect{R}$.
Coordinate transformations are expressed as follows:
\[
    {}^{\B}\vect{p} = {}^{\B}_{\A}\vect{R} {}^{\A}\vect{p} = \vect{C}\big({}^{\B}_{\A}\vect{q}\big){}^{\A}\vect{p}.
\]
We will use the following symbols to distinguish between different types of quantities:
\begin{itemize}
    \setlength\itemsep{0.2em}
    \item a tilde $\widetilde{(\,\cdot\,)}$ means a sensor measurement,
    \item a hat $\widehat{(\,\cdot\,)}$ means an estimate,
    \item a zero superscript $(\,\cdot\,)^0$ means an initial guess (before optimization),
    \item and an asterisk superscript $(\,\cdot\,)^{*}$ means a final value (after optimization).
\end{itemize}

\subsection{Inertial sensor model}

We model the accelerometer measurements from each IMU as
\begin{equation}
\label{eq:accl_measurement}
    ^{I}\widetilde{\vect{a}} = {}^{I}\vect{a}_{\W I} - {}^{I}\vect{g} + \vect{b}_{a} + \vect{n}_{a}
\end{equation}
where
\begin{itemize}
\item ${}^{I}\vect{a}_{\W I}$ is the linear acceleration of the IMU frame with respect to the world frame, expressed in the IMU frame,
\item ${}^{I}\vect{g}$ is the acceleration of gravity expressed in the IMU frame,
\item $\vect{b}_{a}$ is a time-varying bias that is described by a random walk
\[
\vect{b}_{a, k+1} - \vect{b}_{a, k} \sim \sigma_{\vect{b}_a} \sqrt{\Delta t} \cdot \mathcal{N}\left(\vect{0}, \, \vect{1} \right),
\]
\item and $\vect{n}_{a}$ is stochastic noise
\[
\vect{n}_{a} \sim \sigma_{a} / \sqrt{\Delta t} \cdot \mathcal{N}\left(\vect{0}, \, \vect{1} \right)
\]
\end{itemize}
\textcolor{black}{where $\Delta t$ is the sampling interval of the IMU.}
We model the gyroscope measurements from each IMU as 
\begin{equation}
\label{eq:gyro_measurement}
    \editold{^{g}\widetilde{\vect{\omega}}} = \vect{C}(^{g}_{I}\vect{q}) {}^{\I}\vect{\omega}_{\W I} + \vect{b}_{g} + \vect{n}_{g}
\end{equation}
where
\begin{itemize}
\item $\vect{C}(^{g}_{I}\vect{q})$ is the rotation matrix (written in terms of the corresponding quaternion) that describes the orientation of the IMU frame with respect to the gyroscope frame and that allows us to model gyroscope misalignment,
\item ${}^{\I}\vect{\omega}_{\W I}$ is the angular velocity of the IMU frame with respect to the world frame, expressed in the IMU frame,
\item $\vect{b}_{g}$ is a time-varying bias that is described by a random walk
\[
\vect{b}_{g, k+1} - \vect{b}_{g, k} \sim \sigma_{\vect{b}_g} \sqrt{\Delta t} \cdot \mathcal{N}\left(\vect{0}, \, \vect{1} \right),
\]
\item and $\vect{n}_{g}$ is stochastic noise
\[
\vect{n}_{g} \sim \sigma_{g} / \sqrt{\Delta t} \cdot \mathcal{N}\left(\vect{0}, \, \vect{1} \right).
\]
\end{itemize}
In what follows, we assume \textcolor{black}{the square roots of noise density $\sigma_{a}$, $\sigma_{g}$ and bias instability $\sigma_{\vect{b}_a}$, $\sigma_{\vect{b}_g}$ in continuous-time}
have been identified for all sensors prior to extrinsic calibration.

\editold{
We choose not to include the non-orthogonality and scale factors of both the accelerometers and gyroscopes in our calibration process. These parameters can be calibrated in advance using one of several existing approaches~\cite{tedaldi2014robust, sarkka2017multi, ye2017efficient, li2014vector}. We do, on the other hand, model gyroscope misalignment which is frequently ignored in existing calibration processes~\cite{ye2017efficient, li2014vector}. We show in Section~\ref{section:Simulation Experiments} that including these parameters in our calibration process can reduce the RMSE of the remaining calibration parameters. In addition, we choose not to include time offset parameters or asynchronous measurements in our calibration process. There exist several approaches to estimate these parameters separately in software or to correct these issues in hardware \cite{5354093,6782523,9044713}.
}

\subsection{Problem statement (multi-IMU extrinsic calibration)}

Suppose there are $N+1$ IMUs and that we index the frames attached to each of these IMUs as
\[
I_{0}, \, \dotsc, \, I_{N}.
\]
We use 
${}^{I_0}\vect{p}_{I_0{I}_n} := \vect{p}_{{I}_n}$
to denote the relative position of frame $I_{n}$ with respect to frame $I_{0}$, written in the coordinates of frame $I_{0}$.
We use
${}^{I_0}_{I_n}\vect{q} := \vect{q}_{I_n}$
to denote the quaternion describing the relative orientation of frame $I_{n}$ with respect to frame $I_{0}$. 
\editold{Note that these IMU frames $I_{0}, \, \dotsc, \, I_{N}$ are assumed to be aligned with the accelerometers. Gyroscopes measurements are taken in frames $g_{0}, \, \dotsc, \, g_{N}$, where a given frame $g_{n}$ is misaligned from the IMU frame $I_n$ by quaternion $^{g_n}_{I_n}\vect{q}$.}
Given a sequence of measurements
\[
^{I_n}\widetilde{\vect{a}}_{k}, \, \editold{^{g_n}\widetilde{\vect{\omega}}_{k}}
\]
at each time $k \in \{ 1, \dotsc, K \}$ for each IMU $n \in \{ 0, \dotsc, N \},$
the goal of extrinsic calibration is to estimate the parameters
\begin{align*}
& \vect{p}_{{I}_1}, \, \dotsc, \, \vect{p}_{{I}_N} \\
& \vect{q}_{{I}_1}, \, \dotsc, \, \vect{q}_{{I}_N} \\
^{g_{0}}_{I_{0}}\vect{q}, \, &^{g_{1}}_{I_{1}}\vect{q}, \, \dotsc, \, ^{g_{N}}_{I_{N}}\vect{q}.
\end{align*}
To facilitate the estimation of these extrinsic parameters, we have found that it is helpful to simultaneously estimate the time-varying accelerometer and gyroscope biases
\[
\vect{b}_{a_{n}, k}, \, \vect{b}_{g_{n}, k}
\]
at each time $k \in \{ 1, \dotsc, K \}$ for each IMU $n \in \{ 0, \dotsc, N \}$, as well as the angular acceleration of the base IMU
\[
{}^{I_{0}}\vect{\alpha}_{\W I_{0}, k} := ^{I_0}\vect{\alpha}_{k}
\]
at each time $k \in \{ 1, \dotsc, K \}$. These extra parameters can be thought of as auxiliary or slack variables. Note that we do not attempt to estimate the full trajectory of the rigid body to which the IMUs are mounted.

\subsection{Solution approach (maximum-likelihood estimation)}

Defining $\vect{\mathcal{X}}_{0, k} = \{{^{g_0}_{I_0}\vect{q}}, \, {\vect{b}_{a_0, k}},  \, {\vect{b}_{g_0, k}},  \, {}^{I_0}{\vect{\alpha}_{k}} \}$ and \editold{$\vect{\mathcal{X}}_{n, k} = \{ {\vect{p}_{I_n}}, \, {\vect{q}_{I_n}}, \, {^{g_n}_{I_n}\vect{q}}, \, {\vect{b}_{a_n, k}},  \, {\vect{b}_{g_n, k}} \}$}, the overall parameters $\vect{\mathcal{X}} =  \big\{ {\mathcal{X}_{n,k} \, | \, n \in \textcolor{black}{\{0, \dotsc, N \}}, \, k \in \textcolor{black}{\{1, \dotsc, K\}} \big\} }$ can be estimated by solving a non-linear least-square problem
\begin{equation}
\begin{split}
\label{eq:problem statement}
    \min_{\vect{\mathcal{X}}} \Bigg\{ 
    \sum_{ \textcolor{black}{ \substack{n \in \{1, \dotsc, N\} \\ k \in \{1, \dotsc, K\} } } } \Big( & \left\| \vect{r}_{a} \left( \vect{\mathcal{X}}_{n,k}, \; \editold{\vect{\mathcal{X}}_{0,k}}; \; ^{I_n}\widetilde{\vect{a}}_{k},  \; ^{I_0}\widetilde{\vect{a}}_{k}, \; \editold{^{g_0}\widetilde{\vect{\omega}}_{k}} \right) \right\|^{2}_{\vect{\Sigma}_{a}} \\
    + & \left\| \vect{r}_{g} \left( \vect{\mathcal{X}}_{n,k}, \; \editold{\vect{\mathcal{X}}_{0,k}}; \; \editold{^{g_n}\widetilde{\vect{\omega}}_{k}}, \; \editold{^{g_0}\widetilde{\vect{\omega}}_{k}} \right) \right\|^{2}_{\vect{\Sigma}_{g}} \Big) \\
    + \sum_{ \textcolor{black}{ \substack{n \in \{0, \dotsc, N\} \\ k \in \{1, \dotsc, K-1\} } } } \Big( & \left\| \vect{r}_{\vect{b}_a} \left( \vect{\mathcal{X}}_{n,k} \right) \right\|^{2}_{\vect{\Sigma}_{\vect{b}_a}} \\ + & \left\| \vect{r}_{\vect{b}_g} \left( \vect{\mathcal{X}}_{n,k} \right) \right\|^{2}_{\vect{\Sigma}_{\vect{b}_g}} \Big) 
    \Bigg\}
\end{split}
\end{equation}
\editold{where} $ \left\| \cdot \right\|^{2}_{\vect{\Sigma}}$ denotes Mahalanobis distance with a covariance matrix $\vect{\Sigma}$.

The residual $\vect{r}_a$ relates the $n$th accelerometer's measurement ($^{I_n}\widetilde{\vect{a}}_{k}$) compensated by its bias ($\vect{b}_{a_n, k}$) to the specific force reconstructed by the base IMU measurement ($^{I_n}\widehat{\vect{a}}_{k}$)
\begin{equation*}
    \vect{r}_{a} = \left( {}^{I_n}\widetilde{\vect{a}}_{k} - \vect{b}_{a_n, k} \right) - {}^{I_n}\widehat{\vect{a}}_{k}.
\end{equation*}
$^{I_n}\widehat{\vect{a}}_{k}$ is formulated by mapping the specific force at base IMU into the $n$th IMU, parametrized by their geometric relationship and rotational motion
\begin{equation*}
\begin{gathered}
    {}^{I_n}\widehat{\vect{a}}_{k} = {}^{I_0}_{I_n}\vect{R}^{-1} \Big\{ {}^{I_0}\widehat{\vect{a}}_{k} + \left[ ^{I_0}\widehat{\vect{\omega}}_{k} \right]_{\times}^2 \, \vect{p}_{I_n} + [ ^{I_0}\vect{\alpha}_{k}]_{\times} \, \vect{p}_{I_n} \Big\}, \\
    {}^{I_0}\widehat{\vect{a}}_{k} = {}^{I_0}\widetilde{\vect{a}}_{k} - \vect{b}_{a_0, k}, ~ ^{I_0}\widehat{\vect{\omega}}_{k} = {}^{g_0}_{I_0}\vect{R}^{-1} \big( \editold{^{g_0}\widetilde{\vect{\omega}}_{k}} - \vect{b}_{g_0, k} \big)
\end{gathered}
\end{equation*}
where ${}^{I_0}_{I_n}\vect{R} = \vect{C}\big(\vect{q}_{I_n}\big)$, \, ${}^{g_0}_{I_0}\vect{R} = \vect{C}\big(^{g_0}_{I_0}\vect{q}\big)$. Note that $[\cdot]_{\times}$ is an operator converting a cross product into a skew-symmetric matrix form. As the sensor measurement subtracted by their own bias has \editold{an} uncertainty due to zero-mean gaussian white noise---i.e. $\widetilde{\vect{a}} - \vect{b}_a = \vect{n}_a$ from Eq.~(\ref{eq:accl_measurement}) and $\widetilde{\vect{\omega}} - \vect{b}_g = \vect{n}_g$ from Eq.~(\ref{eq:gyro_measurement}), the corresponding covariance matrix for $\vect{r}_a$ is $\vect{\Sigma}_{a} = \{ 2 \cdot {\sigma_{a}}^2 / \Delta t + ({\sigma_{g}}^2 / \Delta t)^{2} \} \cdot \vect{I}_{3 \times 3}$.

$\vect{r}_g$ is a residual relating the $n$th gyroscope's measurement (\editold{$^{g_n}\widetilde{\vect{\omega}}_{k}$}) to that of base gyroscope's (\editold{$^{g_0}\widetilde{\vect{\omega}}_{k}$}), both of which are compensated by their own biases ($\vect{b}_{g_n, k}$, $\vect{b}_{g_0, k}$)
\begin{equation*}
\begin{gathered}
    \vect{r}_{g} = {}^{I_0}\widehat{\vect{\omega}}_{W I_n} - {}^{I_0}\widehat{\vect{\omega}}_{W I_0} \\
    = {}^{I_0}_{I_n}\vect{R} \, {}^{g_n}_{I_n}\vect{R}^{-1} \big( \editold{{}^{g_n}\widetilde{\vect{\omega}}_{k}} - \vect{b}_{g_n, k} \big) - {}^{g_0}_{I_0}\vect{R}^{-1} \big( \editold{{}^{g_0}\widetilde{\vect{\omega}}_{k}} - \vect{b}_{g_0, k} \big)
\end{gathered}
\end{equation*}
where ${}^{I_0}_{I_n}\vect{R} = \vect{C}\big(\vect{q}_{I_n}\big)$, ${}^{g_n}_{I_n}\vect{R} = \vect{C}\big(^{g_n}_{I_n}\vect{q}\big)$, and ${}^{g_0}_{I_0}\vect{R} = \vect{C}\big(^{g_0}_{I_0}\vect{q}\big)$. This relation holds as the same rotational motion is experienced at any locations on the same rigid body. Similar to the covariance matrix for $\vect{r}_{a}$, the covariance matrix for $\vect{r}_{g}$ is $\vect{\Sigma}_{g} = 2 {\sigma_{g}}^2 / \Delta t \cdot \vect{I}_{3 \times 3}$.
\textcolor{black}{Note that $\vect{r}_a$ and $\vect{r}_g$ both take $^{g_0}\widetilde{\vect{\omega}}_{k}$ as input and therefore are correlated.}

$\vect{r}_{\vect{b}_a}$ and $\vect{r}_{\vect{b}_g}$ accounts for bias evolution along time
\begin{gather*}
    \vect{r}_{\vect{b}_a} = \vect{b}_{a_n,k+1} - \vect{b}_{a_n,k} \\
    \vect{r}_{\vect{b}_g} = \vect{b}_{g_n,k+1} - \vect{b}_{g_n,k}.
\end{gather*}
These residuals' covariance matrices $\vect{\Sigma}_{\vect{b}_a}$, $\vect{\Sigma}_{\vect{b}_g}$ comes from the bias evolution in discretized time domain; hence, $\vect{\Sigma}_{\vect{b}_a} = {\sigma_{\vect{b}_a}}^2 \Delta t \cdot \vect{I}_{3 \times 3}$, $\vect{\Sigma}_{\vect{b}_g} = {\sigma_{\vect{b}_g}}^2 \Delta t \cdot \vect{I}_{3 \times 3}$.

\color{black}

\subsection{Degenerate cases}
\label{subsection: Discourse on Degenerate Case}

The solution to Eq.~\eqref{eq:problem statement} can be found when the Fisher information matrix (FIM) is invertible~\cite{jauffret2007observability}. This is identical to the condition when the Jacobian of Eq.~(\ref{eq:problem statement}),
\begin{equation*}
    \vect{J} 
    = 
    \begin{bmatrix}
        \vect{J}_{1, 0} & \vect{J}_{1, 1} & \cdots & \vect{J}_{1, N} \\
        \vect{J}_{2, 0} & \vect{J}_{2, 1} & \cdots & \vect{J}_{2, N} \\
        \vdots          & \vdots          & \ddots & \vdots          \\
        \vect{J}_{K, 0} & \vect{J}_{K, 1} & \cdots & \vect{J}_{K, N} \\
    \end{bmatrix}
\end{equation*}
composed of submatrices at each time $k \in \{1, \dotsc, K\}$ for each IMU $n \in \{0, \dotsc, N\}$
\scriptsize
\begin{equation*}
    \vect{J}_{k, n} = 
    \begin{cases}
    \begin{bmatrix}
        \frac{\partial \vect{r}_a}{\partial \vect{\mathcal{X}}_0} & \frac{\partial \vect{r}_g}{\partial \vect{\mathcal{X}}_0} & \frac{\partial \vect{r}_{b_a}}{\partial \vect{\mathcal{X}}_0} & \frac{\partial \vect{r}_{b_g}}{\partial \vect{\mathcal{X}}_0}
    \end{bmatrix} ^{T}
    \in \mathbb{R}^{12 \times \textcolor{black}{12}} & \text{if $k \neq K, n = 0$} \\
    \begin{bmatrix}
        \frac{\partial \vect{r}_a}{\partial \vect{\mathcal{X}}_n} & \frac{\partial \vect{r}_g}{\partial \vect{\mathcal{X}}_n} & \frac{\partial \vect{r}_{b_a}}{\partial \vect{\mathcal{X}}_n} & \frac{\partial \vect{r}_{b_g}}{\partial \vect{\mathcal{X}}_n}
    \end{bmatrix}^{T}
    \in \mathbb{R}^{12 \times \textcolor{black}{15}} & \text{if $k \neq K, n \neq 0$} \\ 
    \begin{bmatrix}
        \frac{\partial \vect{r}_a}{\partial \vect{\mathcal{X}}_0} & \frac{\partial \vect{r}_g}{\partial \vect{\mathcal{X}}_0}
    \end{bmatrix}^{T}
    \in \mathbb{R}^{6 \times \textcolor{black}{12}} & \text{if $k = K, n = 0$} \\ 
    \begin{bmatrix}
        \frac{\partial \vect{r}_a}{\partial \vect{\mathcal{X}}_n} & \frac{\partial \vect{r}_g}{\partial \vect{\mathcal{X}}_n}
    \end{bmatrix}^{T}
    \in \mathbb{R}^{6 \times \textcolor{black}{15}} & \text{if $k = K, n \neq 0$} \\ 
    \end{cases}
\end{equation*}
\normalsize
is full rank. In this paper, we do not derive all cases where this condition breaks, which we call degenerate cases. However, degenerate cases do exist. The exact set of degenerate cases will vary with respect to the number of IMUs, their geometric arrangement, the number of timesteps, the distribution of measurement, and more.

Previous work on the self-calibration for camera-IMU systems~\cite{kelly2011visual,mirzaei2008kalman} find that motions such as constant acceleration or rotation along a single axis at a constant rate result in a failure mode. In line with this, we run our experiments (wherein we compare our IMU extrinsic calibration techniques to a camera-IMU system) on trajectories where such motions are avoided.

\color{black}

%% file: sections/simulation-experiments.tex
To verify the robustness of our multi-IMU extrinsic calibration \editold{along arbitary trajectories} as stated above, we performed several experiments using OpenVINS~\cite{geneva2020openvins}, an open-source visual-inertial simulator. It is commonly the case that engineering drawings dictate a reasonable initial guess for the (nominal) position and orientation of a set of sensors, but manufacturing tolerances place the sensors at different locations and orientations. Therefore, our robustness evaluation considers sensors placed at a nominal position with initial guesses for parameter estimates selected according to samples from a Gaussian distribution about these nominal positions and orientations. Then we consider the robustness of our solutions where the initial guess for extrinsic parameters may be well outside these tolerances. We show that, even outside typical manufacturing tolerances, our method is robust to different initial conditions.

\subsection{Implementation details}

\begin{figure*}[h!]
    \centering
    \includegraphics[width=\linewidth, keepaspectratio]{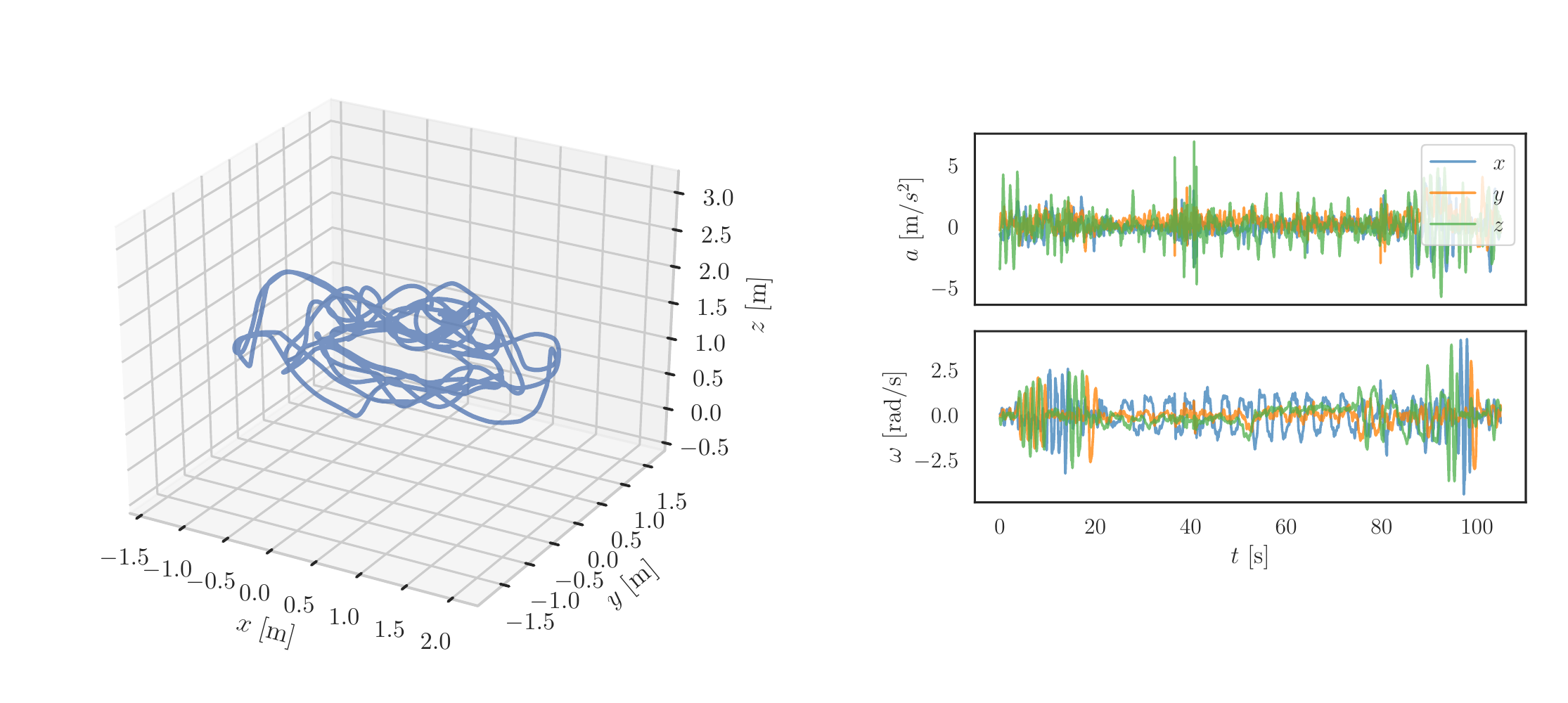}
    \caption{An example trajectory for experiments in simulation (\texttt{room4} in TUM-VI Dataset~\cite{schubert2018vidataset}).}
    \label{fig:tum-vi room4 trajectory}
\end{figure*}

\editold{We assumed that four IMUs were mounted to a rigid body.
Table~\ref{table:sensor layout} shows the reference pose of each IMU.
The accelerometers' noise and random walk characteristics were $\sigma_{a} = 2 \times 10^{-3}~\textrm{m} / \textrm{s}^{2} / \sqrt{\textrm{Hz}}$ and $\sigma_{\vect{b}_a} = 3 \times 10^{-3}~\textrm{m} / \textrm{s}^{2} \cdot \sqrt{\textrm{Hz}}$, while those of gyroscopes were $\sigma_{g} = 1.6968 \times 10^{-4}~\textrm{rad} / \textrm{s} / \sqrt{\textrm{Hz}}$ and $\sigma_{\vect{b}_g} = 1.9393 \times 10^{-5}~\textrm{rad} / \textrm{s} \cdot \sqrt{\textrm{Hz}}$.
The first values of simulated time-varying biases $\vect{b}_{a_{n}, 1}$ and $\vect{b}_{g_{n}, 1}$ were sampled from a uniform distribution on the interval $[-0.05, 0.05]$.
Gyroscope misalignment was generated by rotating the reference orientation of each IMU about an axis chosen uniformly at random by an angle sampled from a zero-mean normal distribution with standard deviation of $1^{\circ}$.
IMU measurements were generated at 100 Hz and have synchronized time stamps while the rigid body was moved along six different trajectories from the TUM-VI Dataset~\cite{schubert2018vidataset}. Fig~\ref{fig:tum-vi room4 trajectory} shows one of these trajectories.}

\begin{table}
    \captionsetup{justification=centering}
    \caption{\editold{\textsc{Reference IMU poses in simulation}}}
    \label{table:sensor layout}
    \begin{center}
    \begin{tabular}{|ccc|}
        \Xhline{\arrayrulewidth}
        index & \editold{position [mm]} & \editold{orientation [rad]*} \\
        \Xhline{\arrayrulewidth}
        IMU0  & [0,   0,   0] & [0, 0, 0] \\
        IMU1  & [200,   0,   0] & [$\pi$, 0, 0] \\
        IMU2  & [0, 200,   0] & [0, $\pi$, 0] \\
        IMU3  & [0,   0, 200] & [0, 0, $\pi$] \\
        \Xhline{\arrayrulewidth}
    \end{tabular}
    \end{center}
    \captionsetup{justification=raggedright}
    \caption*{\footnotesize \editold{\textit{* in this table, orientation is given as XYZ Euler angles}}}
\end{table}

\subsection{Results when initial guesses are within typical manufacturing tolerances}

\editold{
Table~\ref{table:averaged RMSE} shows the root mean square error (RMSE) of the estimated extrinsic parameters---the relative position $\vect{p}_{{I}_n}$ and orientation $\vect{q}_{{I}_n}$ of each IMU $n\in\{1, \dotsc, N\}$ with respect to the base IMU0---over twenty trials along each trajectory in the case when initial guesses were within typical manufacturing tolerances.
The initial guess of each relative position was sampled from a normal distribution with mean equal to the reference position and with a standard deviation along each axis of $5\;\text{mm}$.
The initial guess of each relative orientation was generated by rotating the reference orientation about an axis chosen uniformly at random by an angle sampled from a zero-mean normal distribution with a standard deviation of $5^{\circ}$.
Our results show that, when we included gyroscope misalignment $_{I}^{g}\vect{q}$ in the set of parameters to be estimated (with an initial guess of zero misalignment), our proposed method achieved sub-millimeter and sub-degree error for all trajectories, which is comparable to other existing methods~\cite{kim2017line,schopp2016self}.
Our results also show, in particular, that it is important to estimate gyroscope misalignment (something that not all methods listed in Table~\ref{table:comparison} do, for example \citeauthor{kim2017line}\cite{kim2017line})---without estimating misalignment, position and orientation errors were between one and two orders of magnitude higher.
}

\begin{figure*}[h!]
    \centering
    \includegraphics[width=0.8\linewidth, keepaspectratio]{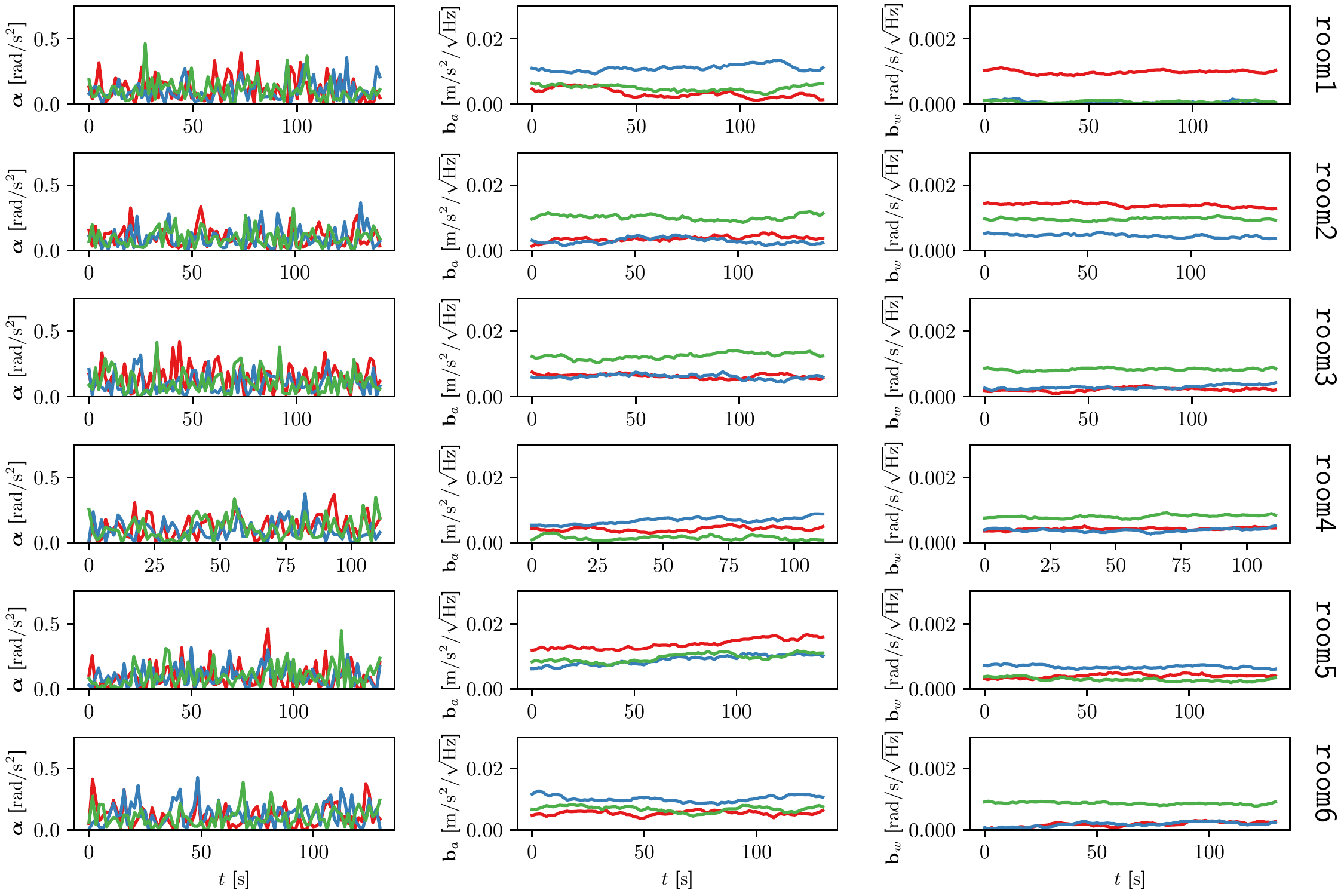}
    \caption{\textcolor{black}{Error of estimated angular acceleration and of each time-varying bias for the base IMU as functions of time over a single trial along six trajectories in simulation (red, green, blue are along x, y, and z axes, respectively).}}
    \label{fig:auxiliary states}
\end{figure*}

\editold{
Table~\ref{table:states RMSE} shows the RMSE of the estimated angular accelerations $^{I_0}\vect{\alpha}_k^{*}$ and time-varying biases $\vect{b}_{a_{n}, k}^{*}$, $\vect{b}_{g_{n}, k}^{*}$ for each time $k \in \{1, \dotsc, K \}$ and each IMU $n \in \{0, \dotsc, N \}$ over the same twenty trials along each trajectory.
The initial guess of each angular acceleration $^{I_0}\vect{\alpha}_k^{0}$ was found by numerical differentiation of gyroscope measurements $^{g_0}\widetilde{\vect{\omega}}_k$ from the base IMU.
The initial guess of each time-varying bias $\vect{b}_{a_{n}, k}^{0}$ and $\vect{b}_{g_{n}, k}^{0}$ was zero.
Recall that angular accelerations and time-varying biases are extra parameters that we think of as auxiliary or slack variables and that their estimation is not our main focus.
Nonetheless, results show that our proposed method reduces RMSE in estimates of all extra parameters when compared to initial guesses.
To help visualize these results,
Fig.~\ref{fig:auxiliary states} shows the \textcolor{black}{error} of each estimated angular acceleration $^{I_0}\vect{\alpha}_{k}^{*}$ and of each time-varying bias $\vect{b}_{a_0, k}^{*}$, $\vect{b}_{g_0, k}^{*}$ for the base IMU as functions of time over a single trial along each trajectory.
}

\renewcommand{\arraystretch}{1.2}
    \begin{table*}
    \captionsetup{justification=centering}
    \caption{\textsc{\editold{RMSE of estimated extrinsic parameters over 20 trials}}}
    \label{table:averaged RMSE}
    \centering
    \begin{tabular}{|ccc|cc|ccc|}
        \Xhline{\arrayrulewidth}
        \multicolumn{3}{|c|}{} & \multicolumn{2}{c|}{\thead{Without Estimation of\\ Gyroscope Misalignment}} & \multicolumn{3}{c|}{\thead{With Estimation of\\ Gyroscope Misalignment}} \\
        trajectory & length [m] & duration [s] & $\vect{p}$ [mm] & $\vect{q}$ [deg] & $\vect{p}$ [mm] & $\vect{q}$ [deg] & $_{I}^{g}\vect{q}$ [deg] \\
        \Xhline{\arrayrulewidth}
        \texttt{room1} & 146.79 & 141.03 & 3.8003 & 1.6397 & 0.1704 & 0.0191 & 0.0391 \\
        \texttt{room2} & 141.60 & 144.07 & 4.3648 & 2.0671 & 0.2318 & 0.0284 & 0.0422 \\
        \texttt{room3} & 135.52 & 141.02 & 3.9474 & 1.1735 & 0.1445 & 0.0167 & 0.0353 \\
        \texttt{room4} & 68.70  & 111.37 & 2.8717 & 1.0320 & 0.1806 & 0.0307 & 0.0651 \\
        \texttt{room5} & 131.64 & 142.32 & 2.5525 & 1.0384 & 0.0809 & 0.0189 & 0.0425 \\
        \texttt{room6} & 67.27  & 130.83 & 4.2959 & 1.1324 & 0.2128 & 0.0150 & 0.0344 \\
        \Xhline{\arrayrulewidth}
    \end{tabular}
    \end{table*}
\renewcommand{\arraystretch}{1.0}

\renewcommand{\arraystretch}{1.5}
\begin{table*}
\captionsetup{justification=centering}
\caption{\editold{\textsc{RMSE of estimated angular accelerations and time-varying biases over 20 trials}}}
\label{table:states RMSE}
\centering
\begin{tabular}{|c|ccc|ccc|}
    \Xhline{\arrayrulewidth}
    \multicolumn{1}{|c|}{} & \multicolumn{3}{c|}{\thead{RMSE of Initial Guesses}} & \multicolumn{3}{c|}{\thead{RMSE of Final Estimates}} \\
    trajectory & $\vect{^{I_0}\alpha}$ [$\textrm{rad} / \textrm{s}^2$]  & $\vect{b}_{a}$ [$\textrm{m} / \textrm{s}^{2} / \sqrt{\textrm{Hz}}$] & $\vect{b}_{g}$  [$\textrm{m} / \textrm{s}^{2} \cdot \sqrt{\textrm{Hz}}$] & $\vect{^{I_0}\alpha}$ [$\textrm{rad} / \textrm{s}^2$]  & $\vect{b}_{a}$ [$\textrm{m} / \textrm{s}^{2} / \sqrt{\textrm{Hz}}$] & $\vect{b}_{g}$  [$\textrm{m} / \textrm{s}^{2} \cdot \sqrt{\textrm{Hz}}$] \\
    \Xhline{\arrayrulewidth}
    \texttt{room1} & 0.3147 & $9.6880 \times 10^{-3}$ & $7.5516 \times 10^{-3}$ & 0.1353 &  $7.2235 \times 10^{-3}$ &  $5.7869 \times 10^{-4}$ \\
    \texttt{room2} & 0.2844 & $1.0285 \times 10^{-2}$ & $9.2693 \times 10^{-3}$ & 0.1256 & $5.9214 \times 10^{-3}$ & $1.0158 \times 10^{-3}$ \\
    \texttt{room3} & 0.2940 & $1.0937 \times 10^{-2}$ & $7.8867 \times 10^{-3}$ & 0.1384 & $9.4185 \times 10^{-3}$ & $5.2697 \times 10^{-4}$ \\
    \texttt{room4} & 0.2360 & $1.0516 \times 10^{-2}$ & $9.3446 \times 10^{-3}$ & 0.1297 & $5.3291 \times 10^{-3}$ & $5.6598 \times 10^{-4}$ \\
    \texttt{room5} & 0.2331 & $1.1235 \times 10^{-2}$ & $9.6848 \times 10^{-3}$ & 0.1288 & $1.0378 \times 10^{-2}$ & $4.9971 \times 10^{-4}$ \\
    \texttt{room6} & 0.2127 & $9.2290 \times 10^{-3}$ & $1.1778 \times 10^{-2}$ & 0.1370 & $7.7067 \times 10^{-3}$ & $5.2553 \times 10^{-4}$ \\
    \Xhline{\arrayrulewidth}
\end{tabular}
\end{table*}

\renewcommand{\arraystretch}{1.0}

\subsection{Results when initial guesses are outside typical manufacturing tolerances}

\renewcommand{\arraystretch}{1.5}
\begin{table}
\captionsetup{justification=centering}
\caption{\editold{\textsc{RMSE of IMU position estimate as a function of difference ($\delta\vect{p}$, $\delta\vect{q}$) between initial guess and reference value \textcolor{black}{(unit: [mm])}}}}
\label{table:average positional error of estimate w.r.t. initial guess deviation}
\centering
\begin{tabular}{|cc|cccc|}
    \Xhline{\arrayrulewidth}
     & & \multicolumn{4}{c|}{$\delta\vect{p}$ [mm]} \\
     & & 0 & 10 & 20 & 30 \\
    \Xhline{\arrayrulewidth}
    \multirow{4}{*}{\rotatebox[origin=c]{90}{$\delta\vect{q}$ [deg]}} &  0 &  0.1767 &  0.1763 &  0.1782 &  0.1765 \\
                                                                      & 30 &  0.1788 &  0.1787 &  0.1777 &  0.1783 \\
                                                                      & 60 &  0.1779 &  0.1773 &  0.1782 &  0.1772 \\
                                                                      & 90 & 39.3668 & 45.8379 & 29.9102 & 60.3195 \\
    \Xhline{\arrayrulewidth}
\end{tabular}
\end{table}
\renewcommand{\arraystretch}{1.0}

\renewcommand{\arraystretch}{1.5}
\begin{table}
\captionsetup{justification=centering}
\caption{\editold{\textsc{RMSE of IMU orientation estimate  as a function of difference ($\delta\vect{p}$, $\delta\vect{q}$) between initial guess and reference value \textcolor{black}{(unit: [deg])}}}}
\label{table:average orientational error of estimate w.r.t. initial guess deviation}
\centering
\begin{tabular}{|cc|cccc|}
    \Xhline{\arrayrulewidth}
     & & \multicolumn{4}{c|}{$\delta\vect{p}$ [mm]} \\
     & & 0 & 10 & 20 & 30 \\
    \Xhline{\arrayrulewidth}
    \multirow{4}{*}{\rotatebox[origin=c]{90}{$\delta\vect{q}$ [deg]}} &  0 & 0.0223 & 0.0221 & 0.0223 & 0.0219 \\
                                                                      & 30 & 0.0226 & 0.0223 & 0.0222 & 0.0222 \\
                                                                      & 60 & 0.0223 & 0.0223 & 0.0223 & 0.0223 \\
                                                                      & 90 & 25.2160 & 5.7559 & 8.3864 & 18.6798 \\
    \Xhline{\arrayrulewidth}
\end{tabular}
\end{table}
\renewcommand{\arraystretch}{1.0}

\color{black}

Next, we study the extent to which the proposed method can reliably estimate extrinsic parameters when initial guesses are outside typical manufacturing tolerances.
    
Table~\ref{table:average positional error of estimate w.r.t. initial guess deviation} shows the RMSE of the estimated position between each IMU $n\in \{1,\dotsc,N\}$ with respect to IMU0 over the same twenty trajectories. Results are reported where the initial guesses for the relative position have between 0 and 30 mm error and initial guesses for relative orientation have between 0 and 90 degrees error. Our results show that---up to 60 degrees of error in the initial orientation and 30 mm of error in the initial relative position---relative position is estimated to within 0.2 millimeters.

Table~\ref{table:average orientational error of estimate w.r.t. initial guess deviation} shows the RMSE of the estimated orientation between each IMU $n\in \{1,\dotsc,N\}$ with respect to IMU0 over the same twenty trajectories. Again, results are reported where the initial guesses for the relative position have between 0 and 30 mm error and initial guesses for relative orientation have between 0 and 90 degrees error. Our results show that---up to 60 degrees of error in the initial orientation and 30 mm of error in the initial relative position---orientation is estimated to within 0.03 degrees.

Again, these initial errors are outside manufacturing tolerances considering the simulated platform's scale as stated in Table~\ref{table:sensor layout}. Therefore, we conclude that our proposed method is robust to a poor initial guess of extrinsic parameters.

%% file: sections/real-world-experiments.tex
\begin{figure}
    \centering
    \includegraphics[width=.4\linewidth, keepaspectratio]{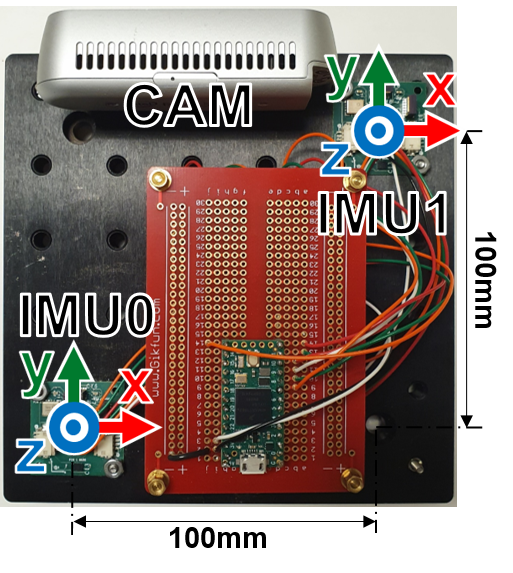}
    \caption{Sensor rig used for hardware experiments.}
    \label{fig:sensor apparatus}
\end{figure}

\color{black}

In this section, we show that the proposed multi-IMU method can estimate extrinsic calibration parameters on par or better than Kalibr, a state-of-the-art and open-source calibration method \cite{rehder2016extending}. In particular, because Kalibr relies upon camera images, it can perform poorly in ill-lit conditions or on trajectories that result in blurry images. The proposed method, because it does not use a camera, performs well despite these conditions. We demonstrate this point by comparing the performance of both methods for trajectories where the calibration image pattern---essential for using Kalibr---generally work well. Next we present calibration results for trajectories where blurry images and ill-lit conditions occur and may cause Kalibr to fail. All data and code are freely available online (see link in Section \ref{section:Introduction}).

\color{black}

\subsection{Implementation details}

\color{black}

Figure~\ref{fig:sensor apparatus} shows the sensor rig that we used to collect a dataset with which to evaluate the performance of our method and of Kalibr. We placed two IMUs (labeled IMU0 and IMU1) and a camera (labeled CAM) on an optical breadboard. Each IMU consisted of one triaxial accelerometer (ST IIS3DWB) and three single-axis gyroscopes (Silicon Sensing CRM100 and CRM200). The camera was an Intel RealSense D435i. We emphasize that this camera was necessary to run Kalibr but was not needed by (and was not used by) our proposed method.

The noise and random walk characteristics of the accelerometers and the gyroscopes were found by analysis of Allan variance in the usual way~\cite{4404126}, and were provided as input both to our method and to Kalibr.
For the accelerometers, we obtained $\sigma_{a} = 1.13 \times 10^{-1}~\textrm{m} / \textrm{s}^{2} / \sqrt{\textrm{Hz}}$ and $\sigma_{\vect{b}_a} = 2.54 \times 10^{-3}~\textrm{m} / \textrm{s}^{2} \cdot \sqrt{\textrm{Hz}}$. For the gyroscopes, we obtained $\sigma_{g} = 3.74 \times 10^{-3}~\textrm{rad} / \textrm{s} / \sqrt{\textrm{Hz}}$ and $\sigma_{\vect{b}_g} = 7.39 \times 10^{-5}~\textrm{rad} / \textrm{s} \cdot \sqrt{\textrm{Hz}}$. IMU measurements were all generated at 100 Hz but were not synchronized.

The intrinsic parameters of the camera (distortion, focal length, and principal point) were found using the camera calibration package from Intel that accompanies the D435i camera, and were provided as input to Kalibr (but not to our method, which does not use camera images).

The intrinsic parameters of each IMU (scale factors and axis nonorthogonality) \textcolor{black}{and the constant time offset} were found by application of Kalibr itself to a preliminary dataset---distinct from the one we subsequently used to test extrinsic calibration---in which the sensor rig moved directly in front of a fiducial marker with excitation along all DOFs, as recommended by the authors of that method~\cite{rehder2016extending}.
There are many other ways in which these intrinsic parameters \textcolor{black}{and the time offset} could have been obtained (and are obtained, in practice)---our choice to use Kalibr itself for intrinsic calibration \textcolor{black}{and time synchronization} was meant to make our subsequent evaluation of extrinsic calibration as favorable to Kalibr as possible, so as to provide a fair comparison with our proposed method.
These intrinsic parameters \textcolor{black}{and the time offset} were provided as input both to our method and to Kalibr.

The reference positions and orientations used to compute errors in estimation of extrinsic calibration parameters were derived from the optical breadboard on which the IMUs were mounted.
In particular, IMU0 and IMU1 were mounted $100\;\text{mm}$ apart along the $x$ and $y$ axes, at the same position along the $z$ axis, and at the same orientation---so, we assumed the following reference values:
\[
\vect{p}_{I_{1}} = \begin{bmatrix} 0.1 & 0.1 & 0.0 \end{bmatrix}^{T}
\qquad
\vect{q}_{I_{1}} = \begin{bmatrix} 0 & 0 & 0 & 1 \end{bmatrix}^{T}
\]
We also assumed a reference value of zero gyroscope misalignment.
We believe these are reasonable assumptions, given that the breadboard has an absolute hole-to-hole tolerance of $\pm 0.01\;\text{mm}$ and a flatness tolerance of $\pm 0.15\;\text{mm}$ over $0.09\;\text{m}^{2}$, and given that PCB manufacturing tolerances are typically sub-millimeter in position and sub-degree in orientation.
However, we acknowledge that the reference positions and orientations we assumed are themselves a source of error in our experiments.
As a consequence, we focus on establishing that both our method and Kalibr produce estimates that have ``comparable'' error with respect to these reference positions and orientations, rather than on rigorously establishing that our method produces ``lower'' error (at least in the baseline condition).

To collect data, we moved and rotated the sensor rig along a number of arbitrary trajectories, each lasting 60 seconds.
To ensure a fair comparison between our method and Kalibr, we kept the calibration pattern in view of the camera during all of these trajectories.
We collected data under three different conditions.
In the {\em baseline} condition (21 trajectories), we used lighting that is known to work well with Kalibr.
In the {\em blurry} condition (23 trajectories), we moved and rotated the rig much more quickly in a way that sometimes resulted in blurry images.
In the {\em ill-lit} condition (21 trajectories), we used poor lighting that makes the calibration pattern less discernible by the Intel D435i camera.

\color{black}

\subsection{Comparison of our method with Kalibr}

\color{black}

Table~\ref{table:IMU Pose Displacement} shows the RMSE of the estimated extrinsic parameters---the relative position $\vect{p}_{{I}_1}$ and orientation $\vect{q}_{{I}_1}$ of IMU1 with respect to the base IMU0---over all trajectories in each condition, obtained both using Kalibr and using our proposed method.
This table also shows the RMSE of gyroscope misalignment (relative to its assumed reference value) when each method is asked to include misalignment---and, in the case of Kalibr, also scale factors and axis non-orthogonality---in the set of parameters to be estimated.
For both methods, the initial guess of relative position was the zero vector and the initial guess of relative orientation was the zero rotation.
For both methods, when gyroscope misalignment was estimated, the initial guess of misalignment was the zero rotation.
For our method, the initial guess of angular acceleration was found by numerical differentiation of gyroscope measurements from IMU0, and the initial guess of each time-varying bias was zero (same as for experiments in simulation).
For both our method and Kalibr under the baseline condition, RMSE in position is on the order of millimeters and RMSE in orientation is on the order of degrees.
The results shown provide evidence that our method matches the performance of Kalibr under the baseline condition regardless of whether or not each method is asked to estimate any other intrinsic parameters.
We do note that, in contrast to simulation, the estimation of gyroscope misalignment (or other intrinsic parameters) seemed to make performance worse both for our method and for Kalibr, although this may be a result of incorrect reference values as discussed in the previous section.

Table~\ref{table:IMU Pose Displacement} also shows the success rate of each method under each condition. We define ``success'' as producing a result, regardless of its quality. For Kalibr, this means completion without any thrown errors. For our method, this means convergence by Ceres Solver~\cite{agarwal2012ceres} without early termination. We define ``success rate'' as the ratio of successes to the total number of trajectories.
The results shown provide evidence that Kalibr often fails in blurry or ill-lit conditions---note, as well, the corresponding increase in RMSE of estimated extrinsic calibration parameters even in those cases when Kalibr succeeds.
Our method is, of course, unaffected by blurry or ill-lit conditions---but the fact that it succeeded across all three sets of trajectories with similar RMSE in all cases does provide evidence that our method is insensitive to the choice of trajectory.

Finally, Table~\ref{table:IMU Pose Displacement} shows the computation time required for each method using an octa-core Intel i7-10700 CPU operating at 2.90GHz with 32GB of RAM. We provide these results to benchmark future work and not to make any claims about the relative efficiency of our method, although we do note that it requires at most about 5 seconds to complete.

\color{black}

\renewcommand{\arraystretch}{1.15}
\begin{table*}
\captionsetup{justification=centering}
\caption{\textsc{\editold{RMSE of estimated extrinsic parameters, computation time, and success rate for both Kalibr and our method when applied to data from hardware experiments}}}
\label{table:IMU Pose Displacement}
\begin{center}
\begin{tabular}{|cc|cc|cc|}
    \Xhline{\arrayrulewidth}
    \multicolumn{2}{|c|}{} & \multicolumn{2}{c|}{\thead{\bf Kalibr}} & \multicolumn{2}{c|}{\thead{\bf Our Method}} \\
    \thead{condition} & & \thead{Without Estimation of\\ Intrinsic Parameters} & \thead{With Estimation of\\ Intrinsic Parameters} & \thead{Without Estimation of\\ Gyroscope Misalignment} & \thead{With Estimation of\\ Gyroscope Misalignment} \\

    \Xhline{\arrayrulewidth}
    \multirow{6}{*}{\textit{baseline}} & RMSE in $\vect{p}$ [mm]          & 3.64 & 4.39 & 1.63 & 1.64 \\
                                       & RMSE in $\vect{q}$ [deg]      & 0.61 & 1.35 & 0.63 & 2.91 \\
                                       & RMSE in $_{I}^{g}\vect{q}$ [deg]      & -    & 2.06 & -    & 2.50 \\
                                       & computation time [ms]  & 56.15 $\pm$ 6.49 & 106.96 $\pm$ 16.34 & 3.62 $\pm$ 0.40 & 4.61 $\pm$ 0.36 \\
                                       & success rate           & 19 / 21 & 19 / 21 & 21 / 21 & 21 / 21 \\
    \Xhline{\arrayrulewidth}
    \multirow{6}{*}{\textit{blurry}} & RMSE in $\vect{p}$ [mm]          & 3.48 & 70.41 & 1.88 & 2.02 \\
                                     & RMSE in $\vect{q}$ [deg]      & 0.69 & 26.98 & 0.68 & 2.86 \\
                                     & RMSE in $_{I}^{g}\vect{q}$ [deg]      & -    & 30.67 & -    & 2.05 \\
                                     & computation time [ms]  & 58.67 $\pm$ 11.36 & 98.99 $\pm$ 12.27 & 4.04 $\pm$ 0.47 & 4.74 $\pm$ 0.58 \\
                                     & success rate           & 13 / 23 & 14 / 23 & 23 / 23 & 23 / 23 \\
    \Xhline{\arrayrulewidth}
    \multirow{6}{*}{\textit{ill-lit}} & RMSE in $\vect{p}$ [mm]          & 119.96 & 107.42 & 1.51 & 1.37 \\
                                      & RMSE in $\vect{q}$ [deg]      & 3.30   & 1.81   & 0.68 & 4.19 \\
                                      & RMSE in $_{I}^{g}\vect{q}$ [deg]      & -      & 1.03   & -    & 3.51 \\
                                      & computation time [ms]  & 8.67   & 7.83 $\pm$ 0.28 & 3.91 $\pm$ 0.44 & 5.16 $\pm$ 0.55 \\
                                      & success rate           & 1 / 21 & 2 / 21 & 21 / 21 & 21 / 21 \\
    \Xhline{\arrayrulewidth}
\end{tabular}
\end{center}
\end{table*}
\renewcommand{\arraystretch}{1.0}

%% file: sections/conclusion.tex
In this paper, we proposed an extrinsic calibration for a multi-IMU system using only measurements \editold{collected along arbitrary trajectories}. We demonstrated the extent to which our method is insensitive to the choice of trajectories through simulations. We also substantiated that the proposed work shows comparable or even exceeds performance compared to a benchmark, which necessitates the use of a camera with a fiducial marker and accompanies its own failure cases. 

\editold{
In this paper, all accelerometer and gyroscope measurements collected along a calibration trajectory are used to compute the optimization cost function. However, it may not be necessary to use all of these measurements in the optimization process. In fact, there exists prior work that does select a subset of measurements for calibration~\cite{schneider2019observability,maye2013self}. Future work along these lines may both improve the computation time of our method and may allow this method to be robust to particular degenerate motions (as we discussed in Section~\ref{subsection: Discourse on Degenerate Case}).
}

\editold{
In this work, we assumed sensors were time synchronized and had the same sample rates. Although there exist methods to synchronize sensors or to subsample the sensors collecting data at different rates, it would be more efficient to incorporate this into our extrinsic calibration process. It may even lead to more accurate extrinsic parameter estimates.
}